\documentclass[final]{cvpr}

\usepackage{times}
\usepackage{epsfig}
\usepackage{graphicx}
\usepackage{amsmath,amssymb}
\usepackage{tabu}
\usepackage{booktabs}
\usepackage{caption}
\usepackage{bm}
\usepackage{xspace}
\usepackage{siunitx}
\usepackage{multirow}
\usepackage{placeins}
\usepackage[numbers,sort]{natbib}
\usepackage{mathtools}

\makeatletter
\newcommand{\settitle}{\@maketitle}
\makeatother

\graphicspath{{figures/}}

\newcommand{\yatagawa}[1]{\textcolor{blue}{\textrm{[Yatagawa: #1]}}}

\DeclareMathOperator*{\argmin}{argmin}

\renewcommand{\etal}{et al.\xspace}

\usepackage[pagebackref=true,breaklinks=true,colorlinks,bookmarks=false]{hyperref}

\usepackage[capitalize]{cleveref}

\def\cvprPaperID{****} 
\def\confYear{CVPR 2021}

\begin{document}

\title{Deep Mesh Prior: Unsupervised Mesh Restoration\\using Graph Convolutional Networks}

\author{Shota Hattori ~~~~~~~ Tatsuya Yatagawa ~~~~~~~ Yutaka Ohtake ~~~~~~~ Hiromasa Suzuki\\
School of Engineering, The University of Tokyo\\
{\tt\small \{hattori,~tatsy,~ohtake,~suzuki\}@den.t.u-tokyo.ac.jp}
}



\maketitle

\begin{abstract}
    This paper addresses mesh restoration problems, i.e., denoising and completion, by learning self-similarity in an unsupervised manner. For this purpose, the proposed method, which we refer to as Deep Mesh Prior, uses a graph convolutional network on meshes to learn the self-similarity. The network takes a single incomplete mesh as input data and directly outputs the reconstructed mesh without being trained using large-scale datasets. Our method does not use any intermediate representations such as an implicit field because the whole process works on a mesh. We demonstrate that our unsupervised method performs equally well or even better than the state-of-the-art methods using large-scale datasets.
\end{abstract}


\section{Introduction}
\label{sec:intro}

\begin{figure}[tbp]
    \centering
    \includegraphics[width=\linewidth]{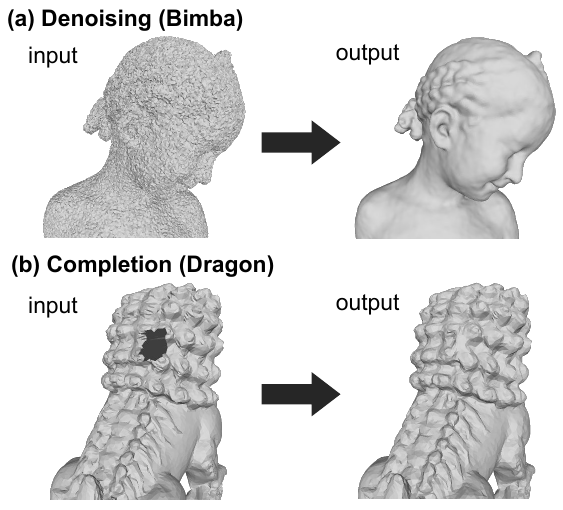}
    \caption{Our method acquires shape priors from a single mesh by unsupervised deep learning. The shape priors can be acquired by the weight sharing architecture of the graph convolutional networks and can be used for mesh restoration tasks such as denoising and completion.}
    \label{fig:abstract}
\end{figure}

Restoring the incompleteness of digital shape data given by 3D scanners is a common problem in computational geometry processing. Noise and deficient regions are among the most typical incompleteness by digital scanners. To address these issues, many approaches for the restoration problems have been proposed~\cite{liepa2003filling,sharf2004coherent,kraevoy2005template,wei2018denoising,li2020dnfnet} that often require prior knowledge to constrain the restored mesh because such problems are often ill-posed. Traditionally, the prior knowledge has been designed based on human heuristics. For completion as an example, the output mesh is often constrained to have local smoothness~\cite{liepa2003filling} and self-similarity~\cite{sharf2004coherent,harary2014context}. Additional examples or templates are often used as prior knowledge to fill the holes~\cite{kraevoy2005template,quinsat2015filling}.

Recent advances in deep learning techniques have delivered shape processing using learning-based prior knowledge typically acquired from large-scale datasets. These approaches have been proposed for various shape representations such as voxels~\cite{varley2017robotic,dai20173depn}, point clouds~\cite{rakotosaona2020pointcleannet,guerrero2018pcpcnet,yuan2018pcn,chen2020unpaired}, and meshes~\cite{dai2019scan2mesh,li2020dnfnet}. However, the current shape datasets used in deep learning include only limited shapes, which causes inappropriate restoration for the shapes not included in the datasets. Unsupervised learning is one of the good solutions to resolve this dependency. Recently, several unsupervised approaches have been proposed, e.g., for semantic segmentation~\cite{chen2019baenet,deng2020cvxnet,chen2020bspnet} and surface reconstruction from dense point clouds~\cite{williams2019dgp,hanocka2020point}. However, unsupervised learning for general-purpose shape restoration remains unchallenged, particularly for meshes.

This paper addresses the reconstruction problems on meshes via unsupervised learning of self-similarity in the input shape. To this end, we extend a previous unsupervised learning method for restoring 2D images~\cite{ulyanov2018deep}, which is known as Deep Image Prior (DIP). Specifically, we re-define the problem for triangular meshes using graph convolutional networks (GCNs)~\cite{kipf2016gcn,defferrard2016chebnet,monti2017geometric}. Analogous to DIP, our method has an appealing aspect of acquiring prior knowledge of the self-similarity only from a single input mesh without being pretrained with any large-scale datasets. Furthermore, both the input and output of our method are a mesh; therefore, our method does not use any intermediate shape representations such as an implicit field.

Most importantly, we found that the straightforward application of the idea of DIP was not the best choice for triangular meshes. Alternatively, we come across an idea to train the neural network to estimate the difference between the input shape and the one abstracted by smoothing. This new pipeline significantly improves the performance of reconstruction tasks, and interestingly, it can achieve comparable or even higher performance than that of the state-of-the-art approaches using deep neural networks and a large-scale training dataset.

\section{Related Work}
\label{sec:related-work}

\subsection{Deep Image Prior and its variants for shapes}

Recently, the nature of neural networks to quickly reconstruct structural information has been found advantageous to restore incompleteness of data only using a single input. DIP~\cite{ulyanov2018deep} is the representative study that has successfully performed restoration tasks such as denoising, super-resolution, and inpainting only using a single corrupted image. It shows that the weight sharing of the convolutional neural network (CNN) is apt to model the repeated and correlated structures. For denoising by DIP, the CNN is trained quickly to reproduce natural images, whereas slowly for noisy ones. For inpainting, the CNN learns the self-similarity from the known image parts and fills the missing regions using the knowledge of self-similarity.

Inspired by DIP, Williams~\etal~\cite{williams2019dgp} proposed Deep Geometric Prior (DGP), a surface reconstruction method from a point cloud, in which several multilayer perceptrons (MLPs) are trained individually to reconstruct different local surface patches. Unfortunately, such individual training without weight sharing cannot capture global self-similarity. Similar to DGP, Hanocka~\etal~\cite{hanocka2020point} proposed Point2Mesh, which reconstructs an entire surface from an input point cloud. Unlike DGP, Point2Mesh uses MeshCNN~\cite{hanocka2019meshcnn} with weight-sharing convolution kernels, which enables to capture the global self-similarity. On the other hand, Point2Mesh reconstructs the surface in a coarse-to-fine fashion. It initially trains a MeshCNN to reproduce the abstracted shape with a specific resolution. To reproduce the more refined shape, it trains another MeshCNN individually. As a result, their method may fail to capture geometric features characterized by different shape resolutions. Thus, the previous methods fail to capture global self-similarity over the entire surface and resolution.

On the other hand, our approach uses a single GCN to capture the self-similarity globally over surface and resolution. Unlike the above methods for surface reconstruction from point clouds, a triangular mesh is used as both the input and output of our method. Therefore, it can be used generally for various geometry processing tasks on meshes. This paper shows the applicability of our method to denoising and completion.

\subsection{Learning-based shape restoration}

Shape restoration is a traditional problem of geometry processing, which involves surface reconstruction~\cite{hoppe1992surface,carr2001reconstruction,ohtake2003multilevel,kazhdan2006poisson,kazhdan2013screened}, denoising~\cite{ohtake2002mesh, Giraudot2013noiseadaptive, wei2018denoising, arvanitis2018fpdenoise}, and shape completion~\cite{liepa2003filling,sharf2004coherent,kraevoy2005template,harary2014context}. Although there are a large number of previous studies for restoration, here, we introduce recent deep learning techniques for different shape representations such as voxels, point clouds, and meshes.

The pioneering technique of 3D shape processing using deep learning is for voxelized shapes~\cite{wu20153d} using volumetric CNNs. Equally, early studies for shape restoration often used voxels as an intermediate representation. For example, incomplete 3D scans are restored by mapping them to a 3D voxel grid and passing them to the volumetric CNNs to obtain the completed shapes~\cite{dai20173depn,varley2017robotic,wu2018learning}. Further, the popular generative adversarial network~\cite{goodfellow2014gan,radford2015unsupervised} is applied to restore voxelized shapes~\cite{wang2017shape,Hermoza2018aogan}. On the other hand, Dai~\etal~\cite{dai2019scan2mesh} converted the input voxels to an undirected graph. Then, a GCN is used on this graph structure to choose the vertices and edges for the output mesh. Unfortunately, these methods for voxelized shapes often suffer from limited output resolution due to their much memory consumption, although there are several memory-efficient network architectures for hierarchical voxel representations~\cite{wang2018aocnn,tatarchenko2017octree}.

Point clouds are another shape representation that has been intensely studied in the area of deep learning. After the groundbreaking neural network architecture of PointNet~\cite{qi2017pointnet,qi2017pointnetpp} was introduced, it has been applied to various restoration tasks such as upsampling~\cite{yu2018pu,Yifan_2019_CVPR}, denoising~\cite{rakotosaona2020pointcleannet}, and completion~\cite{chen2020unpaired,Wen2020skipattention}. A common problem of these studies is that they require reconstructing surfaces from output point clouds to constrain the output surface geometries. A popular workaround is to convert the point cloud to an implicit field defined by a deep neural network~\cite{park2019deepsdf,mescheder2019occupancy,chen2019learning}, which enables to regress the output shapes with the ground truth of the implicit representations. However, such the ground truth data is often hard to obtain, e.g., when the inside and outside of the object are unknown due to the presence of missing parts.

Different from the above two, the mesh representation requires only a relatively small number of vertices and polygons to represent even large and detailed shapes. For its efficiency, the mesh is the standard representation for graphics applications, including rendering and modeling. However, the deep learning techniques for the mesh representation have still been developing, and only a few approaches have been proposed for restoring it. For example, Wang~\etal~\cite{wang2019twostepsdenoise} train two neural networks respectively for filtering noise and recovering details. Another example is DNF-Net~\cite{li2020dnfnet} that corrects noisy surface normals using a deep neural network and recovers a denoised mesh using the surface normals. For completion, Wang~\etal~\cite{wang2020deep} applied the adaptive octree CNN~\cite{wang2018aocnn} (AO-CNN) to restore deficient meshes. However, these approaches use shape datasets to train the neural networks and can be used only for a specific purpose, such as mesh denoising and completion.

Unlike these data-driven approaches, our method is unsupervised. Furthermore, both the input and output of our method are a mesh; hence, it does not depend on any intermediate shape representations.

\section{Deep Mesh Prior}
\label{sec:deep-mesh-prior}

As the restoration problems for 2D images can be defined for pixel values, the restoration problem on meshes, i.e., denoising and completion in this paper, are defined for a function defined over the 2-manifold that may have boundaries. When we apply the DIP's framework directly to a mesh, the function should represent the vertex positions after restoration. However, as we describe in this section, this approach poorly acquires the structural patterns on the mesh because of the different properties of pixel values on the 2D image and vertex positions of the mesh.

\subsection{Problem of DIP for meshes}

DIP restores an input image by mapping another image of random values to the restored image using a CNN. Let an image $I(\bm{x})$ be a function of pixel position $\bm{x}$ and $\sigma(\bm{x})$ be another function giving random values. Then, the output image is represented by $F(\sigma(\bm{x}); \theta)$ where $F$ is a non-linear mapping defined by the CNN with a set of parameters $\theta$. DIP trains the network to solve the following optimization problem:
\begin{align}
    & \theta^{*} = \argmin_\theta \sum_{\bm{x} \in \mathcal{I}} E(\bm{x}), \label{eq:dip-optimization} \\
    & E(\bm{x}) = \| I(\bm{x}) - F(\sigma(\bm{x}); \theta) \|^2, \label{eq:dip-error-norm}
\end{align}
where $\mathcal{I}$ is a set of pixels over the entire image. DIP uses the mean squared error for pixel-wise error function $E$, whereas we can alter it arbitrarily depending on the application.

The restoration problem for a triangular mesh outputs the final vertex positions. Let us define a function $F_\mathcal{M}$ over a 2-manifold $\mathcal{M}$ corresponding to a mesh surface. Here, the 2-manifold $\mathcal{M}$ can be naturally mapped to three-dimensional Euclidean space $\mathbb{R}^3$ by an embedding $\varphi: \mathcal{M} \rightarrow \mathbb{R}^3$. Then, the error in \cref{eq:dip-optimization} can be defined equally for $\varphi$:
\begin{equation}
    E(\bm{x}) = \| \varphi(\bm{x}) - F_\mathcal{M}(\sigma(\bm{x}); \theta) \|^2,
    \label{eq:dmp-simple-optimization}
\end{equation}
where we define $F_\mathcal{M}$ using a GCN on the mesh rather than the CNN for an image.

\begin{figure}[tbp]
    \centering
    \includegraphics[width=\linewidth]{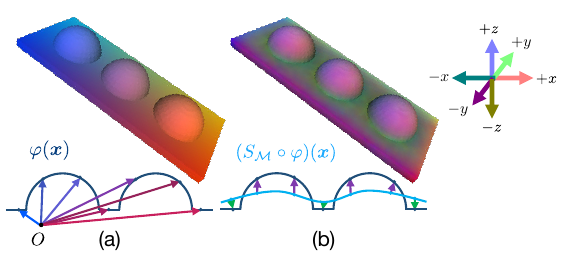}
    \caption{A mesh with repeating hemispherical structures colorized by (a) vertex positions and (b) vertex displacement from a smoothed mesh. The colors correspond to those represented by the orthogonal axes at the right. By subtracting the smoothed mesh from the original one, the repeating structures are extracted as the colors indicate.}
    \label{fig:pos-dis}
\end{figure}

\begin{figure*}[tbp]
    \centering
    \includegraphics[width=\linewidth]{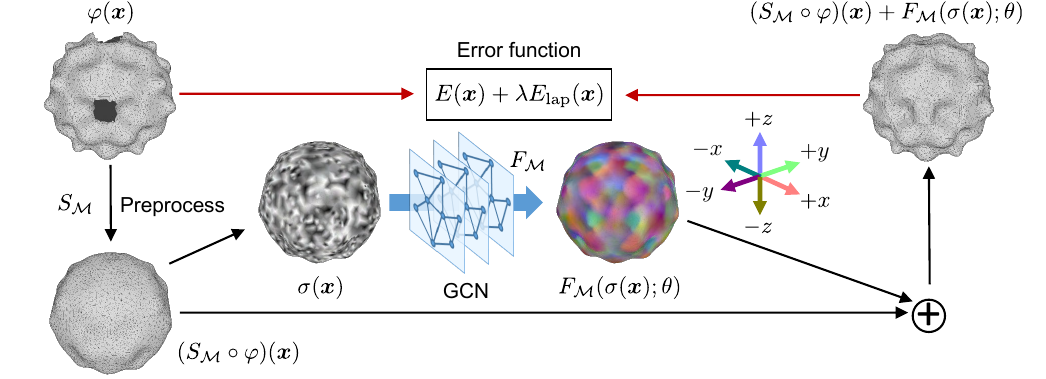}
    \caption{The overview of Deep Mesh Prior for mesh completion. We initially fill the holes of an input deficient mesh so that the mesh has the same topology as an expected output mesh. Then, the Laplacian smoothing is applied to the mesh. The noise function defined over the mesh is mapped to the vertex displacement by a GCN. The displacement is finally added to the smoothed mesh to obtain the output. The GCN is trained to minimize the error between the input and the output meshes.}
    \label{fig:deep_mesh_prior}
\end{figure*}

According to the paper of DIP~\cite{ulyanov2018deep}, its good restoration performance is reasoned by the nature of neural networks more easily learning structural patterns than random ones. However, as we illustrated in \cref{fig:pos-dis}, the vertex positions in $\mathbb{R}^3$ exhibit only a weak structural pattern. Although the color in \cref{fig:pos-dis}(a) changes gradually over the mesh surface, its pattern does not correspond to the geometric structural patterns such as a repetition of hemispherical domes. 


\subsection{Deep priors for vertex displacements}

Assuming that the pixel values defined over a 2D image form a height field, DIP is the process of recovering the height field defined on a flat plane. On the other hand, imagine the earth, which we know is a sphere, a kind of 2-manifold. The terrain on the earth can also be regarded as a height field defined on the sphere. The similarity of these observations suggests that the framework of DIP can work better when applied only to the height component rather than the positions in the 3-dimensional space. We know that the height field for the earth is defined on the sphere, but in general, it is unclear on what geometry the height field is defined. Instead, the height field is defined on an abstracted shape that is given by smoothing the original geometry. Then, we estimate the displacements of the vertices from the smoothed shape in the framework of DIP.

Let $S_\mathcal{M}$ be a filtering operation, then we can write the smoothed vertex positions as $(S_\mathcal{M} \circ \varphi)(\bm{x})$. Figure~\ref{fig:pos-dis}(b) illustrates the displacements between the input positions and smoothed positions. As we can see in this figure, the structural pattern, i.e., the repetition of hemispherical domes, can be well extracted as purple colors by this subtraction.

Now, we train a GCN on the mesh to map random values $\sigma(\bm{x})$ to the vertex displacements that will be added to the smoothed mesh to obtain the output. The reconstruction error in \cref{eq:dmp-simple-optimization} is redefined for the displacement as follows:
\begin{equation}
    E(\bm{x}) = \| \varphi(\bm{x}) - \varphi^{\prime}(\bm{x}) \|,
    \label{eq:dmp-optimization}
\end{equation}
\begin{equation}
    \varphi^{\prime}(\bm{x}) = (S_\mathcal{M} \! \circ \! \varphi)(\bm{x}) + F_\mathcal{M}(\sigma(\bm{x}); \theta).
\end{equation}
Note that our current implementation uses the non-squared Euclidean norm rather than the squared one in \cref{eq:dip-error-norm}. In addition, we minimize the loss function along with the Laplacian loss for mesh vertices:
\begin{equation}
E_\mathrm{lap}(\bm{x}) = \left\| |\mathcal{N}(\bm{x})| \varphi^{\prime}(\bm{x}) -  \sum_{\bm{y} \in \mathcal{N}(\bm{x})} \varphi^{\prime}(\bm{y}) \right\|,
\end{equation}
where $\mathcal{N}(\bm{x})$ is a set of neighboring vertices on the mesh. Finally, the GCN is trained to solve:
\begin{equation}
    \theta^{*} = \argmin_{\theta} \sum_{\bm{x} \in \mathcal{M}} \left( E(\bm{x}) + \lambda E_\mathrm{lap}(\bm{x}) \right),
    \label{eq:dmp-loss-function}
\end{equation}
where $\lambda$ is a parameter to balance the reconstruction error and the Laplacian loss, to which we set different values in different applications.

\begin{figure*}[tbp]
    \centering
    \includegraphics[width=\linewidth]{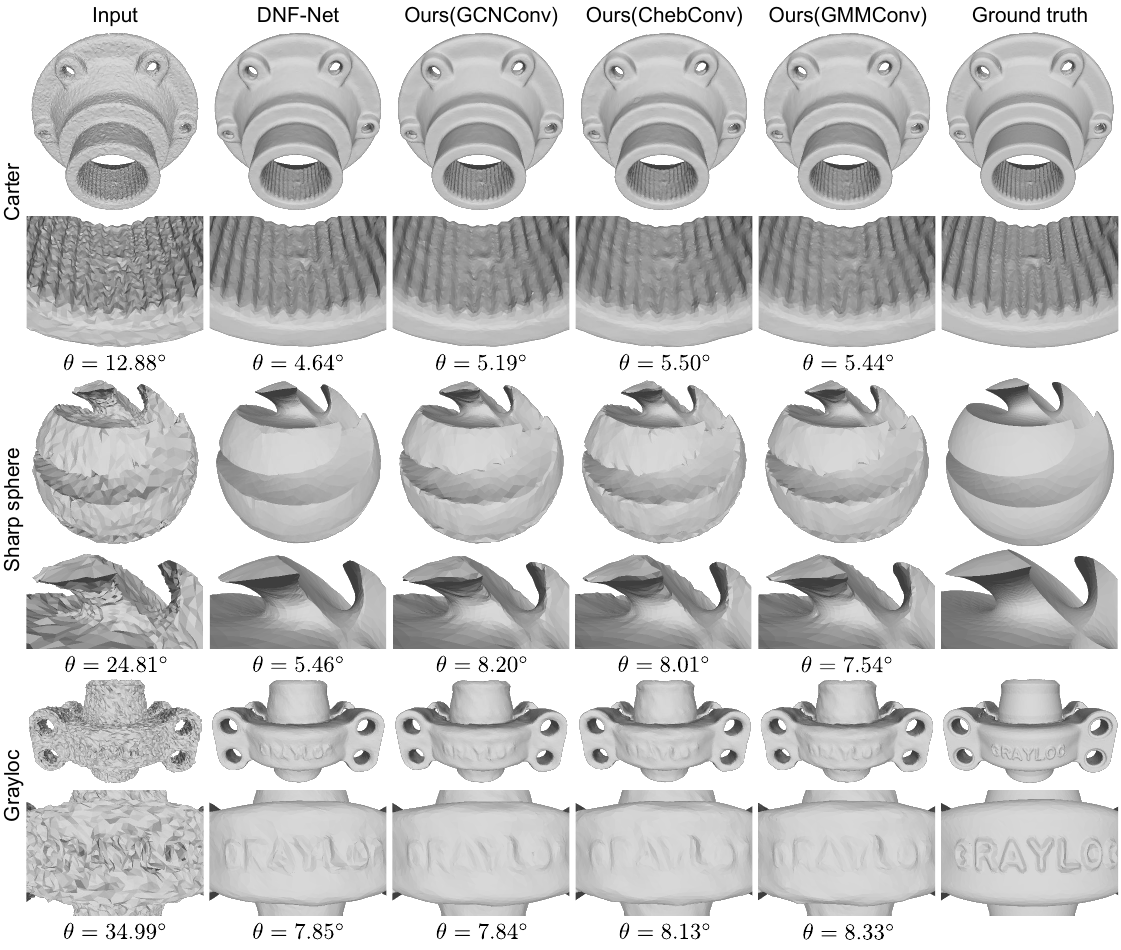}
    \caption{Denoising for three noisy meshes, i.e., ``Carter,'' ``Sharp sphere,'' and ``Grayloc.'' The performance of our method is compared with the state-of-the-art method for denoising (DNF-Net~\cite{li2020dnfnet}) using a large-scale dataset. Despite the independency of the large-scale dataset, our method performs comparably with or even better than the state-of-the-art method.}
    \label{fig:denoising}
\end{figure*}

We refer to this framework as Deep Mesh Prior (DMP) in honor of Deep Image Prior. Figure~\ref{fig:deep_mesh_prior} illustrates the overview of the whole processing pipeline of DMP. This overview depicts an example for mesh completion while our method can also be applied to denoising. To restore the input mesh, we first apply a set of preprocessing and smoothing. Since DMP ensures the same graph structure between the input and output meshes, we need to fill the holes to make the input mesh having the same topology as an expected output mesh. After filling the holes, the vertex positions are smoothed by $S_\mathcal{M}$. Then the noise function $\sigma(\bm{x})$ defined on the mesh manifold is mapped by the GCN to the displacement at each surface position. Finally, we obtain the output mesh by adding the displacements to the vertex positions of the smoothed mesh.

\section{Experiments}
\label{sec:experiments}

We implemented our method using PyTorch~\cite{pytorch} and PyTorch Geometric~\cite{pytorchgeometric}, which is tested on the computer with an NVIDIA GeForce TITAN X (12 GB graphics memory). For smoothing the input mesh, we used the standard Laplacian smoothing without the cotangent weight~\cite{nealen2006laplacian}, which smooths the input mesh such that only the abstract shape remains. The GCN used in this experiment consists of a series of graph-convolutional layers followed by a couple of fully connected layers. For denoising, we tested three different graph convolutional layers, i.e., spectral graph convolution~\cite{kipf2016gcn} (\texttt{GCNConv}), Chebyshev network~\cite{defferrard2016chebnet} (\texttt{ChebConv}), and mixture model network~\cite{monti2017geometric} (\texttt{GMMConv}), while tested only \texttt{GCNConv} for mesh completion. The hyperparameters for each graph convolutional layers are the same as the default values specified by PyTorch Geometric. The network parameters are optimized using Adam~\cite{Kingma2014adam}. We set the learning rate to 0.01 for denoising and 0.001 for mesh completion, while $(\beta_1, \beta_2) = (0.9, 0.999)$ for both of them. For the balancing constant $\lambda$, we used 0.2 for denoising and 0.03 for completion.

\subsection{Mesh denoising}


\begin{figure}[tbp]
    \centering
    \includegraphics[width=\linewidth]{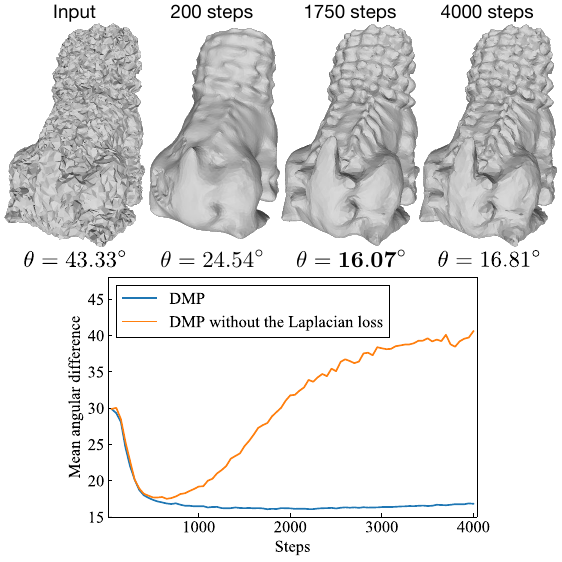}
    \caption{Top images show the results of denoising for the ``Dragon'' model at different training steps. The changes in the MADs for DMP with and without the Laplacian loss are compared in the bottom chart. The Laplacian loss delays the MAD increasing after the local minimum, which allow us to obtain better results easily.}
    \label{fig:ablation_denoise}
\end{figure}

First, we tested the performance of our framework for denoising using three meshes, i.e., Carter, Sharp sphere, and Grayloc. Figure~\ref{fig:denoising} shows these input meshes, corresponding output, and ground truth meshes. In this figure, the performances of the state-of-the-art method, i.e., DNF-Net~\cite{li2020dnfnet}, and our method with different GCNs are qualitatively compared by the appearances of output meshes. They are also compared quantitatively using the mean angular difference (MAD) of face normals against the ground truth shapes. Note that we extracted the best MAD score among those given during 4,000 training steps. As shown in this figure, our method successfully reduces noise, and more surprisingly, its performance is comparable with DNF-Net that uses a large-scale training dataset. On the other hand, these comparisons also imply that our framework works poorly to sharp edges, which can be observed in the results of Sharp sphere. This problem is because our method only focuses on the difference of vertex positions and does not care about the direction of faces. Although a surface shape can be adequately reproduced by regressing the vertex positions when it is smooth, we may need to regress face normals and dihedral angles to reproduce the sharp edges. Even so, our framework is sufficiently valuable, considering that it is free from training with large-scale shape datasets.

Comparing the difference of GCNs used in our framework, their performances did not change significantly for Carter and Grayloc, which are comparatively smooth. In contrast, for Sharp sphere, the performance of \texttt{GMMConv} is better than those of the other two. \texttt{GMMConv} is based on anisotropic Gaussian Mixture Models on the mesh surface, while both \texttt{GCNConv} and \texttt{ChebConv} depend only on the graph structure of the mesh rather than the shape of the mesh. This difference implies \texttt{GMMConv} can be more sensitive to anisotropic geometric features such as the sharp edges of the Sharp sphere model.

When we train the GCN by more training steps than necessary, the network parameters are overfitted to reconstruct noise equally in DIP. Therefore, it is important to know the best step to terminate the training. We found that the best timing for it can be prolonged by using the Laplacian loss in \cref{eq:dmp-loss-function}. Figure~\ref{fig:ablation_denoise} shows the difference of changes in MAD with and without the Laplacian loss. As shown in the bottom chart, the curve without the Laplacian loss has a sharp local minimum, which means that terminating the training at the bottom of the hollow is difficult. In contrast, the curve with the Laplacian loss exhibits a broad and smooth local minimum. As the denoising results in the top also show it, the Laplacian loss allows us to get better results almost independently of the terminating training step.

\subsection{Mesh completion}

Mesh completion using our framework begins with filling the holes by a standard approach~\cite{liepa2003filling} to ensure that the input mesh to the GCN has the same topology with an expected output mesh. After that, we apply Laplacian smoothing to the mesh and pass it to the GCN equally as in denoising. Unlike denoising, we observed that the quality of mesh completion increases when the number of training steps increases. Therefore, for mesh completion, the GCN is trained by at least 8,000 steps until the update of the predicted displacements is sufficiently small. The error $E(\bm{x})$ in \cref{eq:dmp-optimization} is calculated only for the vertices included in the original deficient mesh because the goal positions for inserted vertices to fill the holes are unknown. Even in this way, GCN's weight sharing across the entire mesh works to move the hole-filled regions by learning the self-similarity in an unsupervised manner.

\begin{figure}[tbp]
    \centering
    \includegraphics[width=\linewidth]{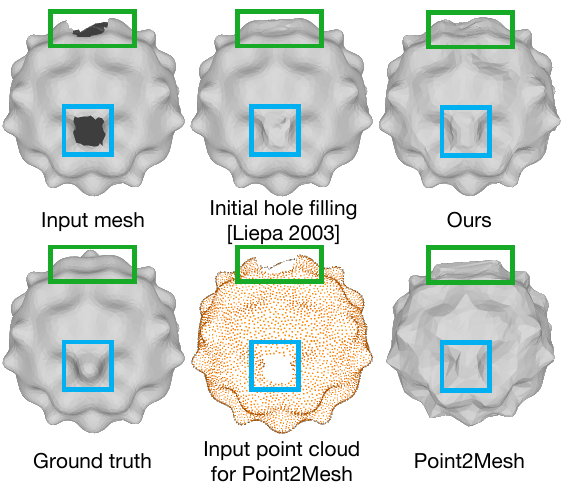}
    \caption{Mesh completion for the ``Bumpy sphere'' model. Although the bumps at regions marked by green and blue rectangles are not reproduced by the initial hole filling~\cite{liepa2003filling} and Point2Mesh~\cite{hanocka2020point}, our method successfully restores the bumps while the entire shape is not the same as the ground truth.}
    \label{fig:completion}
\end{figure}

\begin{figure}[tbp]
    \centering
    \includegraphics[width=\linewidth]{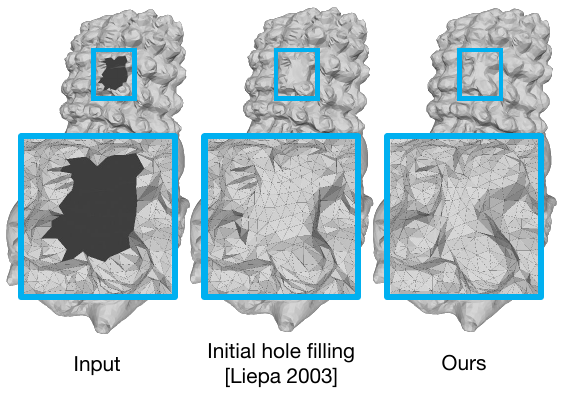}
    \caption{Completion for the ``Dragon'' model. As shown in this figure, the structural shapes on the Dragon's head are enhanced by our method, whereas they are ambiguous only by the initial hole filling.}
    \label{fig:completion_dragon_bimba}
\end{figure}

Figure~\ref{fig:completion} shows the results of mesh completion for a bumpy sphere with two holes. As shown in this figure, the hole regions are flat or noisy after the initial hole filling~\cite{liepa2003filling}. In contrast, the shapes of these regions are enhanced by adding the displacements predicted by the GCN. Then, the bumpy structure of the model is reproduced to be natural. Although the bumpy structure recovered by our method is different from the original mesh at the region highlighted by a green rectangle, the shape itself looks sufficiently natural in terms that two bumps are recovered to follow the repeating bumps at the regions around the hole.

We next compared our method with Point2Mesh~\cite{hanocka2020point}, which is the state-of-the-art surface reconstruction method by learning self-similarity. Since the input for Point2Mesh is a point cloud, we densely sampled 14,534 points on the input deficient mesh as we showed in \cref{fig:completion}. Unfortunately, although Point2Mesh can reconstruct a part of the surface geometry from the dense point cloud, the hole regions with no points remain flat. Furthermore, we found that Point2Mesh requires much more graphics memory than our method because it uses MeshCNN~\cite{hanocka2019meshcnn} to learn the self-similarity that consumes much more memory than standard GCNs. Consequently, the resolution of the output mesh is lower than that of our method when the same amount of graphics memory is available. 

Finally, we tested our method with a more complicated ``Dragon'' model. The result is shown in \cref{fig:completion_dragon_bimba}. As can be seen, the characteristic geometries, such as bumps on the Dragon's head, are ambiguous in the shapes just after the initial hole filling. In contrast, our method enhances these characteristic geometries so that the overall shape of the completed region suits the entire shape of Dragon.

\begin{figure*}[tbp]
    \centering
    \includegraphics[width=\linewidth]{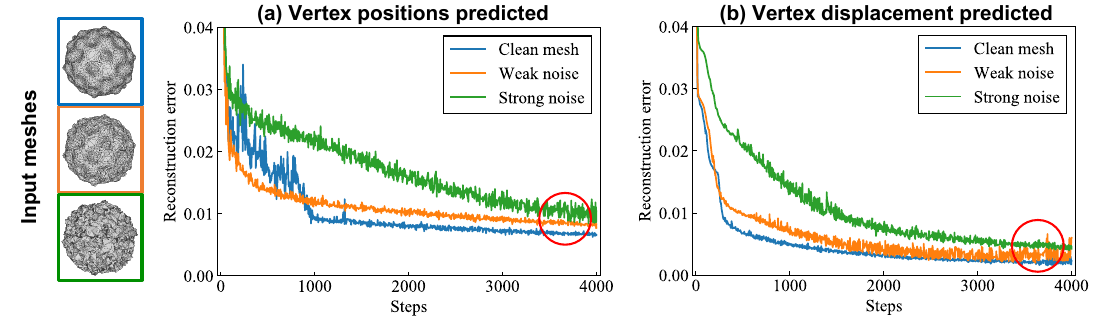}
    \caption{Loss convergence for the reconstruction task using: a clean mesh, that with weak noise, and that with strong noise. We work on the reconstruction task in two ways: (a) directly predicting the vertex positions and (b) predicting the vertex displacements from the smoothed mesh.}
    \label{fig:learning_curves}
\end{figure*}

\begin{table}[tbp]
    \centering
    \begin{tabu} to \linewidth {X[1,l]X[1,c]X[1,c]X[1.2,c]}
        \toprule
        & {Input} & {Pos.} & {Disp. (DMP)} \\
        \cmidrule(r){1-1} \cmidrule{2-4}
        Carter       & 12.88 & 7.85 & \textbf{5.19} \\
        Sharp sph. & 24.81 & 8.25 & \textbf{8.20} \\
        Grayloc      & 34.99 & 9.25 & \textbf{7.84} \\
        \bottomrule
    \end{tabu}
    \caption{Comparison of the MADs for denoising results between the frameworks estimating vertex positions (Pos.) and vertex displacements (Disp.). The latter one for the displacements is equivalent to DMP.}
    \label{tab:pos-dis}
\end{table}

\subsection{Relationship to original DIP}

In DIP for 2D images, the quick fit of the neural network to structured signals is essential to enable its learning framework. Therefore, we checked whether this nature could be observed in our framework for triangular meshes. In this experiment, we prepared three meshes as inputs: a clean mesh and the same one with weak or strong noise. These three types of input mesh are shown at the left of \cref{fig:learning_curves}. As shown in this figure, the original bumpy shape of the input mesh can hardly be observed in the mesh with strong noise, whereas we can see it in that with weak noise. Analogous to DIP, we expect that the neural network can be trained more quickly to reconstruct clean and slightly noisy meshes while slowly to a strongly noisy mesh. In addition, we compare two different approaches (a) to output vertex positions and (b) to output the vertex displacement. The former corresponds to the direct application of DIP to meshes, whereas the latter corresponds to our proposed method. In this experiment, we train the network only using the reconstruction error $E(\bm{x})$ to exclude the influence of the Laplacian loss.

The charts in \cref{fig:learning_curves} show the differences in error convergence. In both approaches (a) and (b), we observed that the convergence rates for strongly noisy meshes (green curve) were significantly slower than those for clean and slightly noisy meshes (blue and orange curves, respectively). This result implies that the network parameters are fitted more quickly to represent structured features as we expected, making it possible to restore meshes using the framework of DIP. Comparing two approaches (a) and (b), we found that the error values of (b) were smaller than those of (a), particularly at the later stage of training.

The above difference of the error values can be observed in the results of denoising. Two strategies of predicting (a) vertex positions and (b) vertex displacements are compared in the MAD for three example meshes shown in \cref{fig:denoising}. The MAD scores are shown in \cref{tab:pos-dis}. The difference in the MAD scores between the two strategies implies the difficulty of predicting vertex positions as structured signals by the GCN. Thus, our method of predicting the vertex displacements is better suited to the shape restoration tasks.

\begin{figure}[tbp]
    \centering
    \includegraphics[width=\linewidth]{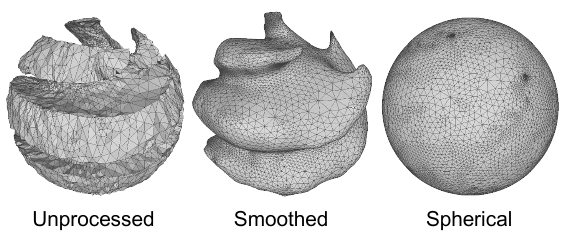}
    \caption{Different mesh appearances obtained by three mesh smoothing operations compared in \cref{tab:smooth_spherical}.}
    \label{fig:preprocessing}
\end{figure}

\begin{table}[tbp]
    \centering
    \begin{tabu} to \linewidth {X[1,l]X[1,c]X[1,c]X[1,c]}
        \toprule
        & {Unprocessed} & {Smoothed} & {Spherical}\\
        \cmidrule(r){1-1} \cmidrule{2-4}
        Sharp sph. & 24.80 & \textbf{8.20} & 8.59\\
        Dragon & 43.26 & \textbf{16.07} & 16.07\\
        Bimba & 37.09 & \textbf{5.17} & 6.91\\
        \bottomrule
    \end{tabu}
    \caption{Comparison of the MADs for denoising results obtained by different initial mesh smoothing operations.}
    \label{tab:smooth_spherical}
\end{table}

\subsection{Effect of smoothing}

In our method, the smoothing operation to the input mesh plays an important role in extracting the vertex displacements. To understand the effect of smoothing, we compared the Laplacian smoothing, which we used in the above experiment, with two other options, i.e., not processing the input and mapping to the spherical surface~\cite{gotsman2003fundamental}. Figure~\ref{fig:preprocessing} shows the meshes obtained by these three operations. In this experiment, we used ``Dragon,'' and ``Bimba'' models instead of ``Carter'' and ``Grayloc'' in \cref{fig:denoising} because the latter two are not homeomorphic to a sphere.

Table~\ref{tab:smooth_spherical} shows the comparison using the MAD given by the different smoothing operations. As shown in this table, when the input mesh is not smoothed, the input and output for the GCN are the same. Then, the GCN is immediately trained to output zero displacements. Therefore, the MAD score for the non-smoothed mesh is approximately the same as that for the noisy input mesh. On the other hand, when the input mesh is initially mapped to a spherical surface, we can obtain the MAD score close to that given by the Laplacian smoothing. However, the significant difference of the sphere and output shape delays the convergence of training. For example, for Bimba model, the best MAD score by 4,000 training steps is 6.91 as in \cref{tab:smooth_spherical} while it improves to 5.82 by 10,000 training steps. According to these observations, the smoothness of the input for the GCN is an important factor for our method, whereas over-smoothing causes slow convergence and results in a worse performance. Thus, the Laplacian smoothing is the most practical choice for the initial mesh smoothing.

\section{Conclusion}
\label{sec:conclusion}

This paper presents an unsupervised learning method for mesh restoration problems, which leverages the prior knowledge given by the GCN. We refer to the proposed method as DMP, which directly predicts a reconstruction mesh from only a single input mesh, including noise and deficient regions. Since DMP is based on the framework of DIP, it does not require pretraining and any large-scale shape dataset. Also, since DMP performs the entire process on a mesh, it does not require any intermediate shape representations to get the final output mesh. The experimental results shown in this paper demonstrate that the performance of DMP is comparable with or even higher than the state-of-the-art learning-based method using a large-scale dataset. Since our framework is sufficiently general to other shape representations, we would like to investigate the applicability of DMP to such representations for future work.

\section*{Acknowledgment}

This work is financially supported by the JSPS Grand-in-Aid for Young Researchers (JP18K18075).




{\small
\bibliographystyle{ieee_fullname}
\bibliography{egbib}

\begin{thebibliography}{10}\itemsep=-1pt

\bibitem{arvanitis2018fpdenoise}
Gerasimos Arvanitis, Aris Lalos, Konstantinos Moustakas, and Nikos Fakotakis.
\newblock Feature preserving mesh denoising based on graph spectral processing.
\newblock {\em IEEE Transactions on Visualization and Computer Graphics},
  25:1513 -- 1527, 2018.

\bibitem{carr2001reconstruction}
J.~C. Carr, R.~K. Beatson, J.~B. Cherrie, T.~J. Mitchell, W.~R. Fright, B.~C.
  McCallum, and T.~R. Evans.
\newblock Reconstruction and representation of 3d objects with radial basis
  functions.
\newblock In {\em Proceedings of ACM International Conference on Computer
  Graphics and Interactive Techniques (SIGGRAPH)}, page 67–76, 2001.

\bibitem{chen2020unpaired}
Xuelin Chen, Baoquan Chen, and Niloy~J Mitra.
\newblock Unpaired point cloud completion on real scans using adversarial
  training.
\newblock {\em Proceedings of International Conference on Learning
  Representations (ICLR)}, 2020.

\bibitem{chen2020bspnet}
Zhiqin Chen, Andrea Tagliasacchi, and Hao Zhang.
\newblock Bsp-net: Generating compact meshes via binary space partitioning.
\newblock In {\em Proceedings of IEEE Conference on Computer Vision and Pattern
  Recognition (CVPR)}, pages 45--54, 2020.

\bibitem{chen2019baenet}
Zhiqin Chen, Kangxue Yin, Matthew Fisher, Siddhartha Chaudhuri, and Hao Zhang.
\newblock Bae-net: Branched autoencoder for shape co-segmentation.
\newblock In {\em Proceedings of IEEE Conference on Computer Vision and Pattern
  Recognition (CVPR)}, pages 8490--8499, 2019.

\bibitem{chen2019learning}
Zhiqin Chen and Hao Zhang.
\newblock Learning implicit fields for generative shape modeling.
\newblock In {\em Proceedings of IEEE Conference on Computer Vision and Pattern
  Recognition (CVPR)}, pages 5939--5948, 2019.

\bibitem{dai2019scan2mesh}
Angela Dai and Matthias NieBner.
\newblock Scan2mesh: From unstructured range scans to 3d meshes.
\newblock In {\em Proceedings of IEEE Conference on Computer Vision and Pattern
  Recognition (CVPR)}, pages 5569--5578, 2019.

\bibitem{dai20173depn}
Angela Dai, Charles Ruizhongtai~Qi, and Matthias NieBner.
\newblock Shape completion using 3d-encoder-predictor cnns and shape synthesis.
\newblock In {\em Proceedings of IEEE Conference on Computer Vision and Pattern
  Recognition (CVPR)}, pages 6545--6554, 2017.

\bibitem{defferrard2016chebnet}
Micha{\"e}l Defferrard, Xavier Bresson, and Pierre Vandergheynst.
\newblock Convolutional neural networks on graphs with fast localized spectral
  filtering.
\newblock In {\em Advances in Neural Information Processing Systems (NIPS)},
  page 3844–3852, 2016.

\bibitem{deng2020cvxnet}
Boyang Deng, Kyle Genova, Soroosh Yazdani, Sofien Bouaziz, Geoffrey Hinton, and
  Andrea Tagliasacchi.
\newblock Cvxnet: Learnable convex decomposition.
\newblock In {\em Proceedings of IEEE Conference on Computer Vision and Pattern
  Recognition (CVPR)}, pages 31--44, 2020.

\bibitem{pytorchgeometric}
Matthias Fey and Jan~E. Lenssen.
\newblock Fast graph representation learning with {PyTorch Geometric}.
\newblock In {\em ICLR Workshop on Representation Learning on Graphs and
  Manifolds}, 2019.

\bibitem{Giraudot2013noiseadaptive}
Simon Giraudot, David Cohen-Steiner, and Pierre Alliez.
\newblock Noise-adaptive shape reconstruction from raw point sets.
\newblock In {\em Proceedings of Eurographics Symposium on Geometry Processing
  (SGP)}, page 229–238, 2013.

\bibitem{goodfellow2014gan}
Ian Goodfellow, Jean Pouget-Abadie, Mehdi Mirza, Bing Xu, David Warde-Farley,
  Sherjil Ozair, Aaron Courville, and Yoshua Bengio.
\newblock Generative adversarial nets.
\newblock In Z. Ghahramani, M. Welling, C. Cortes, N. Lawrence, and K.~Q.
  Weinberger, editors, {\em Advances in Neural Information Processing Systems
  (NIPS)}, volume~27, 2014.

\bibitem{gotsman2003fundamental}
Craig Gotsman, Xianfeng Gu, and Alla Sheffer.
\newblock Fundamentals of spherical parameterization for 3d meshes.
\newblock In {\em Proceedings of ACM International Conference on Computer
  Graphics and Interactive Techniques (SIGGRAPH)}, page 358–363, 2003.

\bibitem{guerrero2018pcpcnet}
Paul Guerrero, Yanir Kleiman, Maks Ovsjanikov, and Niloy~J. Mitra.
\newblock {PCPNet}: Learning local shape properties from raw point clouds.
\newblock {\em Computer Graphics Forum}, 37(2):75--85, 2018.

\bibitem{hanocka2019meshcnn}
Rana Hanocka, Amir Hertz, Noa Fish, Raja Giryes, Shachar Fleishman, and Daniel
  Cohen-Or.
\newblock Meshcnn: A network with an edge.
\newblock {\em ACM Transactions on Graphics}, 38(4), 2019.

\bibitem{hanocka2020point}
Rana Hanocka, Gal Metzer, Raja Giryes, and Daniel Cohen-Or.
\newblock Point2mesh: A self-prior for deformable meshes.
\newblock {\em ACM Transactions on Graphics}, 39(4), 2020.

\bibitem{harary2014context}
Gur Harary, Ayellet Tal, and Eitan Grinspun.
\newblock Context-based coherent surface completion.
\newblock {\em ACM Transactions on Graphics}, 33(1), 2014.

\bibitem{Hermoza2018aogan}
Renato Hermoza and Ivan Sipiran.
\newblock 3d reconstruction of incomplete archaeological objects using a
  generative adversarial network.
\newblock In {\em Proceeding of Computer Graphics International}, pages 5--11,
  2018.

\bibitem{hoppe1992surface}
Hugues Hoppe, Tony DeRose, Tom Duchamp, John McDonald, and Werner Stuetzle.
\newblock Surface reconstruction from unorganized points.
\newblock In {\em Proceedings of ACM International Conference on Computer
  Graphics and Interactive Techniques (SIGGRAPH)}, page 71–78, 1992.

\bibitem{kazhdan2006poisson}
Michael Kazhdan, Matthew Bolitho, and Hugues Hoppe.
\newblock Poisson surface reconstruction.
\newblock In {\em Proceedings of Eurographics Symposium on Geometry Processing
  (SGP)}, volume~7, 2006.

\bibitem{kazhdan2013screened}
Michael Kazhdan and Hugues Hoppe.
\newblock Screened poisson surface reconstruction.
\newblock {\em ACM Transactions on Graphics}, 32(3), 2013.

\bibitem{Kingma2014adam}
Diederik~P. Kingma and Jimmy Ba.
\newblock Adam: {A} method for stochastic optimization.
\newblock In Yoshua Bengio and Yann LeCun, editors, {\em Proceedings of
  International Conference on Learning Representations (ICLR)}, 2015.

\bibitem{kipf2016gcn}
Thomas~N Kipf and Max Welling.
\newblock Semi-supervised classification with graph convolutional networks.
\newblock In {\em Proceedings of International Conference on Learning
  Representations (ICLR)}, 2017.

\bibitem{kraevoy2005template}
Vladislav Kraevoy and Alla Sheffer.
\newblock Template-based mesh completion.
\newblock In {\em Proceedings of Eurographics Symposium on Geometry Processing
  (SGP)}, page 13–22, 2005.

\bibitem{li2020dnfnet}
Xianzhi Li, Ruihui Li, Lei Zhu, Chi-Wing Fu, and Pheng-Ann Heng.
\newblock Dnf-net: a deep normal filtering network for mesh denoising.
\newblock {\em IEEE Transactions on Visualization and Computer Graphics}, 2020.

\bibitem{liepa2003filling}
Peter Liepa.
\newblock {Filling Holes in Meshes}.
\newblock In {\em Proceedings of Eurographics Symposium on Geometry Processing
  (SGP)}, 2003.

\bibitem{mescheder2019occupancy}
Lars Mescheder, Michael Oechsle, Michael Niemeyer, Sebastian Nowozin, and
  Andreas Geiger.
\newblock Occupancy networks: Learning 3d reconstruction in function space.
\newblock In {\em Proceedings of IEEE Conference on Computer Vision and Pattern
  Recognition (CVPR)}, pages 4460--4470, 2019.

\bibitem{monti2017geometric}
Federico Monti, Davide Boscaini, Jonathan Masci, Emanuele Rodola, Jan Svoboda,
  and Michael~M Bronstein.
\newblock Geometric deep learning on graphs and manifolds using mixture model
  cnns.
\newblock In {\em Proceedings of IEEE Conference on Computer Vision and Pattern
  Recognition (CVPR)}, pages 5115--5124, 2017.

\bibitem{nealen2006laplacian}
Andrew Nealen, Takeo Igarashi, Olga Sorkine, and Marc Alexa.
\newblock Laplacian mesh optimization.
\newblock In {\em Proceedings of the International Conference on Computer
  Graphics and Interactive Techniques in Australasia and Southeast Asia
  (GRAPHITE)}, page 381–389, 2006.

\bibitem{ohtake2003multilevel}
Yutaka Ohtake, Alexander Belyaev, Marc Alexa, Greg Turk, and Hans-Peter Seidel.
\newblock Multi-level partition of unity implicits.
\newblock {\em ACM Transactions on Graphics}, 22(3):463–470, 2003.

\bibitem{ohtake2002mesh}
Yutaka Ohtake, Alexander~G Belyaev, and Hans-Peter Seidel.
\newblock Mesh smoothing by adaptive and anisotropic gaussian filter applied to
  mesh normals.
\newblock In {\em Proceeding of Vision, Modeling, and Visuallization (VMV)},
  volume~2, pages 203--210. Citeseer, 2002.

\bibitem{park2019deepsdf}
Jeong~Joon Park, Peter Florence, Julian Straub, Richard Newcombe, and Steven
  Lovegrove.
\newblock Deepsdf: Learning continuous signed distance functions for shape
  representation.
\newblock In {\em Proceedings of IEEE Conference on Computer Vision and Pattern
  Recognition (CVPR)}, pages 165--174, 2019.

\bibitem{pytorch}
Adam Paszke, Sam Gross, Francisco Massa, Adam Lerer, James Bradbury, Gregory
  Chanan, Trevor Killeen, Zeming Lin, Natalia Gimelshein, Luca Antiga, Alban
  Desmaison, Andreas Kopf, Edward Yang, Zachary DeVito, Martin Raison, Alykhan
  Tejani, Sasank Chilamkurthy, Benoit Steiner, Lu Fang, Junjie Bai, and Soumith
  Chintala.
\newblock Pytorch: An imperative style, high-performance deep learning library.
\newblock In {\em Advances in Neural Information Processing Systems (NIPS)},
  pages 8024--8035. 2019.

\bibitem{qi2017pointnet}
Charles~R Qi, Hao Su, Kaichun Mo, and Leonidas~J Guibas.
\newblock Pointnet: Deep learning on point sets for 3d classification and
  segmentation.
\newblock In {\em Proceedings of IEEE Conference on Computer Vision and Pattern
  Recognition (CVPR)}, pages 652--660, 2017.

\bibitem{qi2017pointnetpp}
Charles~R Qi, Li Yi, Hao Su, and Leonidas~J Guibas.
\newblock Pointnet++: Deep hierarchical feature learning on point sets in a
  metric space.
\newblock In {\em Advances in Neural Information Processing Systems (NIPS)},
  2017.

\bibitem{quinsat2015filling}
Yann Quinsat et~al.
\newblock Filling holes in digitized point cloud using a morphing-based
  approach to preserve volume characteristics.
\newblock {\em International Journal of Advanced Manufacturing Technology},
  81(1):411--421, 2015.

\bibitem{radford2015unsupervised}
Alec Radford, Luke Metz, and Soumith Chintala.
\newblock Unsupervised representation learning with deep convolutional
  generative adversarial networks.
\newblock {\em arXiv preprint arXiv:1511.06434}, 2015.

\bibitem{rakotosaona2020pointcleannet}
Marie-Julie Rakotosaona, Vittorio La~Barbera, Paul Guerrero, Niloy~J Mitra, and
  Maks Ovsjanikov.
\newblock Pointcleannet: Learning to denoise and remove outliers from dense
  point clouds.
\newblock {\em Computer Graphics Forum}, 39(1):185--203, 2020.

\bibitem{sharf2004coherent}
Andrei Sharf, Marc Alexa, and Daniel Cohen-Or.
\newblock Context-based surface completion.
\newblock {\em ACM Transactions on Graphics}, 23(3):878–887, 2004.

\bibitem{tatarchenko2017octree}
Maxim Tatarchenko, Alexey Dosovitskiy, and Thomas Brox.
\newblock Octree generating networks: Efficient convolutional architectures for
  high-resolution 3d outputs.
\newblock In {\em Proceedings of International Conference on Computer Vision
  (ICCV)}, pages 2088--2096, 2017.

\bibitem{ulyanov2018deep}
Dmitry Ulyanov, Andrea Vedaldi, and Victor Lempitsky.
\newblock Deep image prior.
\newblock In {\em Proceedings of IEEE Conference on Computer Vision and Pattern
  Recognition (CVPR)}, pages 9446--9454, 2018.

\bibitem{varley2017robotic}
Jacob Varley, Chad DeChant, Adam Richardson, Joaquín Ruales, and Peter Allen.
\newblock Shape completion enabled robotic grasping.
\newblock In {\em Proceedings of IEEE/RSJ International Conference on
  Intelligent Robots and Systems (IROS)}, pages 2442--2447, 2017.

\bibitem{wang2019twostepsdenoise}
Jun Wang, Jin Huang, Fu~Lee Wang, Mingqiang Wei, Haoran Xie, and Jing Qin.
\newblock Data-driven geometry-recovering mesh denoising.
\newblock {\em Computer-Aided Design}, 114:133--142, 2019.

\bibitem{wang2020deep}
Peng-Shuai Wang, Yang Liu, and Xin Tong.
\newblock Deep octree-based cnns with output-guided skip connections for 3d
  shape and scene completion.
\newblock In {\em Proceedings of IEEE Conference on Computer Vision and Pattern
  Recognition (CVPR)}, pages 266--267, 2020.

\bibitem{wang2018aocnn}
Peng-Shuai Wang, Chun-Yu Sun, Yang Liu, and Xin Tong.
\newblock Adaptive o-cnn: A patch-based deep representation of 3d shapes.
\newblock {\em ACM Transactions on Graphics}, 37(6), 2018.

\bibitem{wang2017shape}
Weiyue Wang, Qiangui Huang, Suya You, Chao Yang, and Ulrich Neumann.
\newblock Shape inpainting using 3d generative adversarial network and
  recurrent convolutional networks.
\newblock In {\em Proceedings of International Conference on Computer Vision
  (ICCV)}, 2017.

\bibitem{wei2018denoising}
Mingqiang Wei, Jin Huang, Xingyu Xie, Ligang Liu, Jun Wang, and Jing Qin.
\newblock Mesh denoising guided by patch normal co-filtering via kernel
  low-rank recovery.
\newblock {\em IEEE Transactions on Visualization and Computer Graphics},
  25(10):2910--2926, 2019.

\bibitem{Wen2020skipattention}
Xin Wen, Tianyang Li, Zhizhong Han, and Yu-Shen Liu.
\newblock Point cloud completion by skip-attention network with hierarchical
  folding.
\newblock In {\em Proceedings of IEEE Conference on Computer Vision and Pattern
  Recognition (CVPR)}, 2020.

\bibitem{williams2019dgp}
Francis Williams, Teseo Schneider, Claudio Silva, Denis Zorin, Joan Bruna, and
  Daniele Panozzo.
\newblock Deep geometric prior for surface reconstruction.
\newblock In {\em Proceedings of IEEE Conference on Computer Vision and Pattern
  Recognition (CVPR)}, pages 10122--10131, 2019.

\bibitem{wu2018learning}
Jiajun Wu, Chengkai Zhang, Xiuming Zhang, Zhoutong Zhang, William~T. Freeman,
  and Joshua~B. Tenenbaum.
\newblock Learning shape priors for single-view 3d completion and
  reconstruction.
\newblock In {\em Proceedings of the European Conference on Computer Vision
  (ECCV)}, 2018.

\bibitem{wu20153d}
Zhirong Wu, Shuran Song, Aditya Khosla, Fisher Yu, Linguang Zhang, Xiaoou Tang,
  and Jianxiong Xiao.
\newblock 3d shapenets: A deep representation for volumetric shapes.
\newblock In {\em Proceedings of IEEE Conference on Computer Vision and Pattern
  Recognition (CVPR)}, pages 1912--1920, 2015.

\bibitem{Yifan_2019_CVPR}
Wang Yifan, Shihao Wu, Hui Huang, Daniel Cohen-Or, and Olga Sorkine-Hornung.
\newblock Patch-based progressive 3d point set upsampling.
\newblock In {\em Proceedings of IEEE Conference on Computer Vision and Pattern
  Recognition (CVPR)}, 2019.

\bibitem{yu2018pu}
Lequan Yu, Xianzhi Li, Chi-Wing Fu, Daniel Cohen-Or, and Pheng-Ann Heng.
\newblock Pu-net: Point cloud upsampling network.
\newblock In {\em Proceedings of IEEE Conference on Computer Vision and Pattern
  Recognition (CVPR)}, 2018.

\bibitem{yuan2018pcn}
Wentao Yuan, Tejas Khot, David Held, Christoph Mertz, and Martial Hebert.
\newblock Pcn: Point completion network.
\newblock In {\em Proceedings of IEEE International Conference on 3D Vision
  (3DV)}, pages 728--737, 2018.

\end{thebibliography}
}

\end{document}